\documentclass[english]{cccconf}
\usepackage[comma,numbers,square,sort&compress]{natbib}
\usepackage{amsfonts,amsmath,amssymb,amsthm}
\usepackage{mathrsfs}
\usepackage{subfig}
\usepackage{bm}
\newcommand{\vl}{\;|\;}
\begin{document}

\title{Automatic Identification of Coal and Rock/Gangue Based on DenseNet and Gaussian Process}

\author{Yufan Li\aref{1}}



\affiliation[1]{Department of Automation, Shanghai Jiao Tong University, Shanghai 200240, P.~R.~China
\email{yufanli@sjtu.edu.cn}}

\maketitle

\begin{abstract}
To improve the purity of coal and prevent damage to the coal mining machine, it is necessary to identify coal and rock in underground coal mines. At the same time, the mined coal needs to be purified to remove rock and gangue. These two procedures are manually operated by workers in most coal mines. The realization of automatic identification and purification is not only conducive to the automation of coal mines, but also ensures the safety of workers. We discuss the possibility of using image-based methods to distinguish them. In order to find a solution that can be used in both scenarios, a model that forwards image feature extracted by DenseNet to Gaussian process is proposed, which is trained on images taken on surface and achieves high accuracy on images taken underground. This indicates our method is powerful in few-shot learning such as identification of coal and rock/gangue and might be beneficial for realizing automation in coal mines.
\end{abstract}

\keywords{Mining, Coal and rock, Automatic identification, DenseNet, Gaussian process}


\section{Introduction}
Coal is an important resource in China. With the increasing demand for coal resources, coal in the shallow layers of the earth has been depleted and mining of deep coal mines becomes mainstream. However, excessive coal mining has led to the instability of the crustal rock shell, which intensifies the risk of manual down-hole operation, such as coal seam flake gang, gas explosion and other geological disasters. An important reason for the frequent occurrence of accidents in Chinese coal industry is the low level of automation and intelligence of coal mining equipment, as well as lack of labor-intensive condition under coal mines. Therefore, improving the level of automation of coal mining equipment and improving the labor conditions of workers will enhance the operational safety of the coal industry.

In actual operation, the cutting state of the coal mining machine changes significantly when cutting into the rock, and the overall operating condition of the coal mining machine will be drastically affected by the change. In order to ensure the smooth and reliable operation of the coal mining machine, it is necessary to regulate the coal mining machine according to the distribution of coal and rock in the coal-rock interface of the comprehensive mining working face. Improving the purity of the mined coal is another challenge. In the process of coal collection, the purity of collected coal varies. It is mixed with some gangue, which is a solid waste with low carbon and high sulfur content, and will produce toxic gases such as sulfur dioxide when burned. This reduces the quality of coal and is not conducive to environmental protection. Therefore, the identification of coal and rock/gangue becomes an urgent problem to be solved.

In this paper, we propose an architecture which achieves high accuracy and effectiveness on identification of coal and rock/gangue: use DenseNet \cite{huang2017densely} to extract features, and forward them to Gaussian process in order to classify. Although DenseNet is a model that is not prone to overfitting, using a single DenseNet does not work well due to the presence of noise in the labels. Classifying by Gaussian process only with feature set as raw pixel RGB values is another choice, but it usually results in a huge amount of computation, and it is common practice to use convolutional neural networks (CNNs) to extract features and downsample. We also observe that manual feature extraction does not perform well on our datasets. So we discard the classification head of DenseNet and classify its last hidden state directly using Gaussian process. Experiments show that the combination of the two is very beneficial for describing the distribution of coal/rock/gangue images even in the presence of noise in the labels and performs better than existing image-based methods.
\section{Related Work}
In the past 20 years, researchers have been looking for a general and easy-to-use method for automatic identification of coal and rock/gangue. Early surveys \cite{sun2011study,ge2016internet} pointed out that there are two main streams of current coal rock identification: sensor-based and image-based.

\textbf{Sensor-based methods}. Sensor-based methods include $\gamma$ ray detection, radar detection, infrared temperature measurement, laser detection and vibration spectrum, etc. Bessinger et al. establish a coal rock stratification thickness model using $\gamma$ ray \cite{bessinger1993remnant}. Wang et al. prove that calculation of radar echo wave reflected from the coal rock interface can be used to estimate the thickness of the coal seam \cite{wang2016effect}. Ralston et al. present a sensing method to automatically track geological coal seam features based on thermal infrared imaging \cite{ralston2013developing}. Furthermore, Wang et al. use thermal image to identify coal-rock interface dynamically \cite{wang2019dynamic}. Similarly, a way to classify coal and gangue based on thermal image is proposed in \cite{alfarzaeai2020coal}. The vibration characteristics of coal cutter can also be useful \cite{jie2016recognition}. Shao et al. set up a calibration-free method by Hyperspectral LiDAR to classify coal and rock \cite{shao201991,zixin2021classification}. Li et al. propose a method which detects the sound hitting the central slot by wavelet transform and fast Fourier transform to measure the coal-rock interface based on the principle that the sound falling on the central slot of the scraper conveyor is different due to the different density of coal rock \cite{li2009new,tao2009new}.

\textbf{Image-based methods}. Image-based methods extract image features to distinguish coal and rock/gangue. \cite{liu2000automatic} can be treated as the earliest work using image-based method. \cite{wu2017coal,hu2019multispectral,liu2019multi} implement classification of coal and rock using local binary pattern (LBP) feature \cite{ojala1996comparative}, when \cite{haonan2010research} extracts the gray-level co-occurrence matrix (GLCM) \cite{haralick1973textural} features of coal and gangue, \cite{wang2020lbp} combines GLCM and LBP features of coal/rock images to make up for the shortcomings of a single LBP method. Multi-scale wavelet statistics (MSWS) \cite{de1997non} is another feature which can be used to classify \cite{sun2013image,sun2013wave}. Xue performs dimensionality reduction of images using principal component analysis (PCA) then classify it directly with support vector machine (SVM) \cite{xue2022coal}.

In recent years, with the growth of computer computing power and the development of deep learning, automatic feature extraction is simpler and more general compared to hand-designed features, and some work on coal/rock/gangue pictures using neural networks has also emerged. Wang et al. discover that CNN can distinguish the two even when the gangue surface is covered with coal dust or the surface of the coal block has more mirror tissue \cite{wang2019cnn}. Pu et al. apply transfer learning by using pre-training weights of VGG-16 \cite{pu2019image}. In \cite{si2020deep}, Si et al. implement classification of coal image, coal-rock mixed image and rock image.

Although they demonstrate that it is possible to distinguish with images, those methods suffer from a number of problems. First, the majority of the work uses relatively small data sets that are prone to overfitting, and the high test accuracy may also be due to chance factors. Second, most of images they use are taken in bright surface environments and they do not use those taken under low-light coal shafts for verification. While DenseNet reduces over-fitting on tasks with smaller training set sizes, Gaussian process is also suitable for few-shot and interfered sample learning \cite{wang2021learning}. We also test our model on real images taken under coal mine and achieve high accuracy. Generally, our method can learn coal/rock/gangue images distribution as much as possible with a small datasets and work better with some anti-interference capability.

\textit{Note:} Because rock and gangue are both unwanted contents during mineral, for simplicity, we will refer them collectively as \textbf{coal waste} and coal/rock/gangue images as \textbf{coal images} subsequently.
\section{DenseNet-GP}
\begin{figure*}[!t]
\centering
\includegraphics[width=7.in]{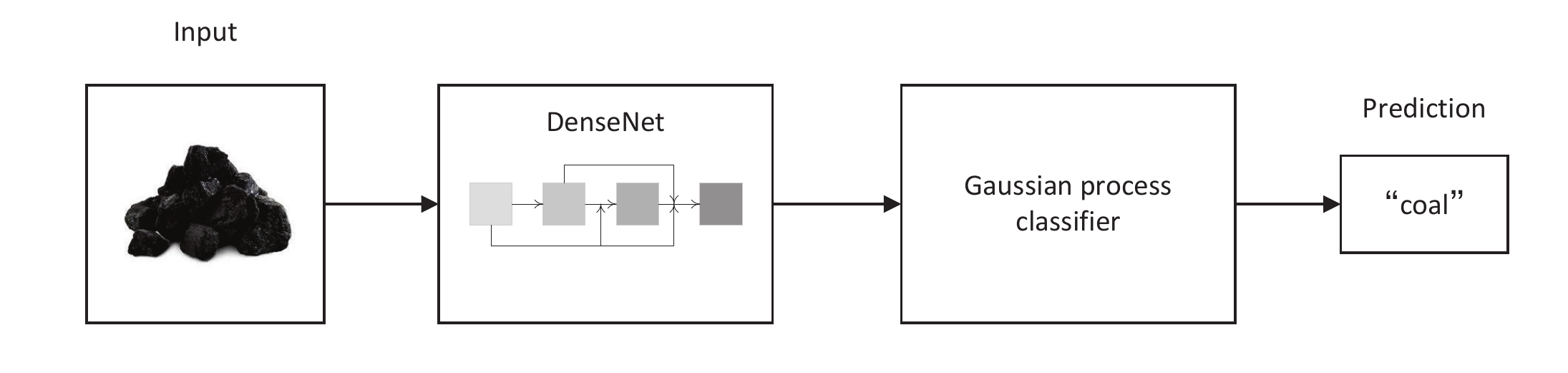}
\caption{Architecture of our model. Each small square in DenseNet represents a convolutional layer, and each layer takes as input the output of all the layers that precede it.}
\label{model}
\end{figure*}
The proposed model can be summarized as the combination of DenseNet and Gaussian process, as shown in Fig.~\ref{model}.
\subsection{DenseNet}\label{dense}
Densely connected convolutional networks (DenseNet) is an improved CNN \cite{huang2017densely}. Instead of summing inputs and outputs in each layer \cite{he2016deep}, DenseNet combines them by concatenation, and each layer takes all outputs from preceding blocks as its input rather than only the previous one, which leads to an implicit deep supervision \cite{lee2015deeply}.

In regular CNNs, the input of $i^\mathrm{th}$ layer is a non-linear mapping $\mathcal{F}_i(\cdot)$ of $(i-1)^\mathrm{th}$ layer:
\begin{equation}
\mathbf x_i = \mathcal F_i(\mathbf x_{i-1}).
\end{equation}
In original ResNet, the input of $i^\mathrm{th}$ layer is the summation of the input and output of $(i-1)^\mathrm{th}$ layer:
\begin{equation}
\mathbf x_i = \mathcal F_i(\mathbf x_{i-1})+\mathbf x_{i-1}.
\end{equation}
And in DenseNet, the input of $i^\mathrm{th}$ layer is the concatenation of outputs from all preceding layers:
\begin{equation}
\mathbf x_i = \mathcal F_i([\mathbf x_0,\mathbf x_1,\cdots,\mathbf x_{i-1}]).
\end{equation}
The major part of DenseNet are dense blocks and transition layers. Each dense block is made up of an $1\times 1$ and a $3\times 3$ convolutional layer repeated $N$ times, while the transition layer consists of an $1\times 1$ convolutional layer and a $2\times 2$ average pooling layer with stride 2.

While ResNet makes it possible to train networks with more than a hundred layers, its huge number of parameters results its low efficiency. Huang et al. find that not all layers in deep residual networks are needed \cite{huang2016deep}. Unlike the later improved ResNets \cite{he2016identity,szegedy2017inception}, which benefit from deep or wide architectures, DenseNet uses feature reuse to gather information. Hence, gradients can be passed throughout the entire network, and a smaller number of layers is demanded to train. Due to these properties of DenseNet, it is a good choice to use it as a feature extractor for coal rock identification. Subsequent experiments corroborate its powerful feature description capability.
\subsection{Gaussian process}

Consider a Gaussian process (GP) given by
\begin{equation}
f(\mathbf X)\sim\mathcal N(\mu(\mathbf X),\mathbf K),
\end{equation}
where $\mathbf K=\kappa(\mathbf X,\mathbf X)$ is a parameteric kernel function. If all of training data are labeled rightly, we have $\sigma_Y^2=0$. However, in our identify task, there might be some noise in training images.

In Gaussian process for regression, we want to predict $f(\mathbf X_*)$ at new points $\mathbf X_*$ based on noisy values $\mathbf Y$ observed at points $\mathbf X$. The joint distribution below obviously satisfies the infinite dimensional Gaussian distribution:
\begin{equation}
\label{joint gp}
\begin{bmatrix}
\mathbf Y \\
f(\mathbf X_*)
\end{bmatrix}=\begin{bmatrix}
\mathbf f \\
\mathbf f_*
\end{bmatrix}\sim\mathcal N\left(
\begin{bmatrix}
\bm\mu \\
\bm\mu_*
\end{bmatrix},\begin{bmatrix}
\mathbf K_Y & \mathbf k_*\\
\mathbf k_*^\top & \mathbf k_{**}
\end{bmatrix}\right),
\end{equation}
where $\mathbf K_Y=\mathbf K+\sigma^2_Y\mathbf I$, $\mathbf k_*=\kappa(\mathbf X,\mathbf X_*)$, $\mathbf k_{**}=\kappa(\mathbf X_*,\mathbf X_*)$. The following conditional distribution is also another Gaussian process\footnote{The formula for conditioning a joint Gaussian distribution is 
\begin{align*}
\begin{bmatrix}
	\mathbf x \\
	\mathbf y
\end{bmatrix}&\sim\mathcal N\left(\begin{bmatrix}
	\mathbf a \\
	\mathbf b
\end{bmatrix},\begin{bmatrix}
	\mathbf A & \mathbf C\\
	\mathbf C^{\begin{small}\top\end{small}} & \mathbf B
\end{bmatrix}\right)\Rightarrow\\\mathbf x\vl \mathbf y&\sim\mathcal N(\mathbf C\mathbf B^{-1}(\mathbf y-\mathbf b)+\mathbf a,\mathbf A-\mathbf C\mathbf B^{-1}\mathbf C^\top).
\end{align*}
}:
\begin{equation}
\label{condition gp}
\mathbf f_*\vl \mathbf f\sim\mathcal N(\mathbf k_*^\top \mathbf K^{-1}(\mathbf f-\bm\mu)+\bm\mu_*,\mathbf k_{**}-\mathbf k_*^\top \mathbf K^{-1}\mathbf k_*),
\end{equation}
which is the posterior distribution.

When it comes to binary classification, our goal is to calculate the probability $p(t_*=1\vl \mathbf t)$ based on observed targets $\mathbf t$. Denote $p(t=1\vl a)=\sigma(a)$, the predictive distribution can be defined as
\begin{equation}
\begin{aligned}
p(t_*=1\vl\mathbf t)&=\int p(t_*=1\vl a_*)p(a_*\vl \mathbf t)\, da_*\\
&=\int \sigma(a_*)p(a_*\vl \mathbf t)\, da_*.
\end{aligned}
\end{equation}
We can approximate $p(a_*\vl \mathbf t)$ with a Gaussian distribution
\begin{equation}
\label{pa*|t}
p(a_*\vl \mathbf t)=\int p(a_*\vl\mathbf a)p(\mathbf a\vl\mathbf t)\,d\mathbf a,
\end{equation}
where $p(a_*\vl\mathbf a)$ can be obtained by \eqref{condition gp}. The Taylor expansion of $\ln\mathbf a$ at point $\hat{\mathbf a}$ is
\begin{equation}
\ln \mathbf a\simeq\ln \hat{\mathbf a}-\frac{1}{2}(\mathbf a-\hat{\mathbf a})^\top\mathbf A(\mathbf a-\hat{\mathbf a}),
\end{equation}
where
\begin{equation*}
	\mathbf A=-\nabla\nabla\ln \mathbf a\Big|_{\mathbf a=\hat{\mathbf a}}=\mathbf K_a^{-1}=(\mathbf K+\sigma_a^2\mathbf I)^{-1}
\end{equation*}
is the $M\times M$ Hessian matrix. There might be some noise in our image labels, and the difference in distribution of pictures on the surface and under the coal shaft can also be regarded as noise. Hence the posterior distribution $p(\mathbf a\vl\mathbf t)$ can be approximated with a Gaussian distribution using the Laplace approximation \cite{bishop2006pattern}
\begin{equation}
\begin{aligned}
q(\mathbf a)&=\frac{(\det\mathbf H)^{1/2}}{(2\pi)^{M/2}}\exp\left\{-\frac{1}{2}(\mathbf a-\hat{\mathbf a})^\top\mathbf H(\mathbf a-\hat{\mathbf a})\right\}\\
&\sim\mathcal N(\hat{\mathbf a},\mathbf{H}^{-1}),
\end{aligned}
\end{equation}
where $\mathbf H=\mathbf W+\mathbf A$, and
\begin{equation}
\begin{aligned}
\mathbf W&=\begin{bmatrix}
	\sigma(a_1)(1-\sigma(a_1)) & & \\
	& \ddots & \\
	& & \sigma(a_M)(1-\sigma(a_M))
\end{bmatrix}\\
&=\bm\sigma(\mathbf 1-\bm\sigma).
\end{aligned}
\end{equation}
The iterative equations is
\begin{equation*}
\mathbf a'=\mathbf K_a(\mathbf I+\mathbf W\mathbf K_a)^{-1}(\mathbf t-\bm\sigma+\mathbf{Wa})
\end{equation*}
and $\mathbf a'$ converges to $\hat{\mathbf a}$. As two components in \eqref{pa*|t} are both Gaussian, its result is also a Gaussian given by\footnote{It follows that $X\sim\mathcal N(\mu,\sigma)\Leftrightarrow p(X)=\mathcal N(X\vl\mu,\sigma)$ and\\ $X\vl Y\sim\mathcal N(\mu,\sigma)\Leftrightarrow p(X\vl Y)=\mathcal N(X\vl\mu,\sigma).$}
\begin{equation}
p(a_*\vl \mathbf t)\approx\mathcal N(a_*\vl\mu_{a_*},\sigma^2_{a_*}),
\end{equation}
where
\begin{equation*}
\begin{aligned}
\mu_{a_*}&=\mathbf k_*^\top(\mathbf t-\bm\sigma),\\
\sigma^2_{a_*}&=\mathbf k_{**}-\mathbf k_*^\top(\mathbf W^{-1}+\mathbf K_a)^{-1}\mathbf k_{*}.
\end{aligned}
\end{equation*}
Finally we obtain the predictive distribution\footnote{$\displaystyle\int\sigma(a)\mathcal N(a\vl\mu,\sigma^2)\simeq\sigma\left(\kappa(\sigma^2)\mu\right)$, see \cite{bishop2006pattern}.}
\begin{equation}
p(t_*=1\vl\mathbf t)\approx\sigma\left(\kappa(\sigma^2_{a_*})\mu_{a_*}\right).
\end{equation}
\section{Experiments}
We will show effectiveness of DenseNet-GP model on our datasets compared with existing popular methods, including SVM, Vision Transformer, ResNet and DenseNet-only model.
\subsection{Datasets}
\begin{figure*}[!htbp]
	\centering
	\subfloat[]{\includegraphics[width=1.7in,height=1.5in]{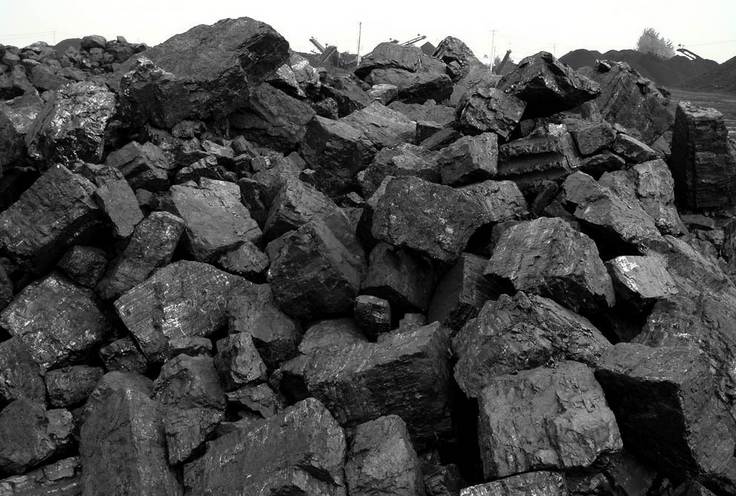}}
	\hfil
	\subfloat[]{\includegraphics[width=1.7in,height=1.5in]{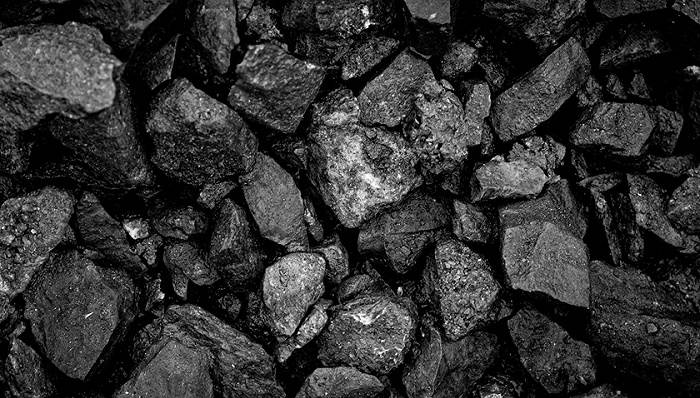}}
	\hfil
	\subfloat[]{\includegraphics[width=1.7in,height=1.5in]{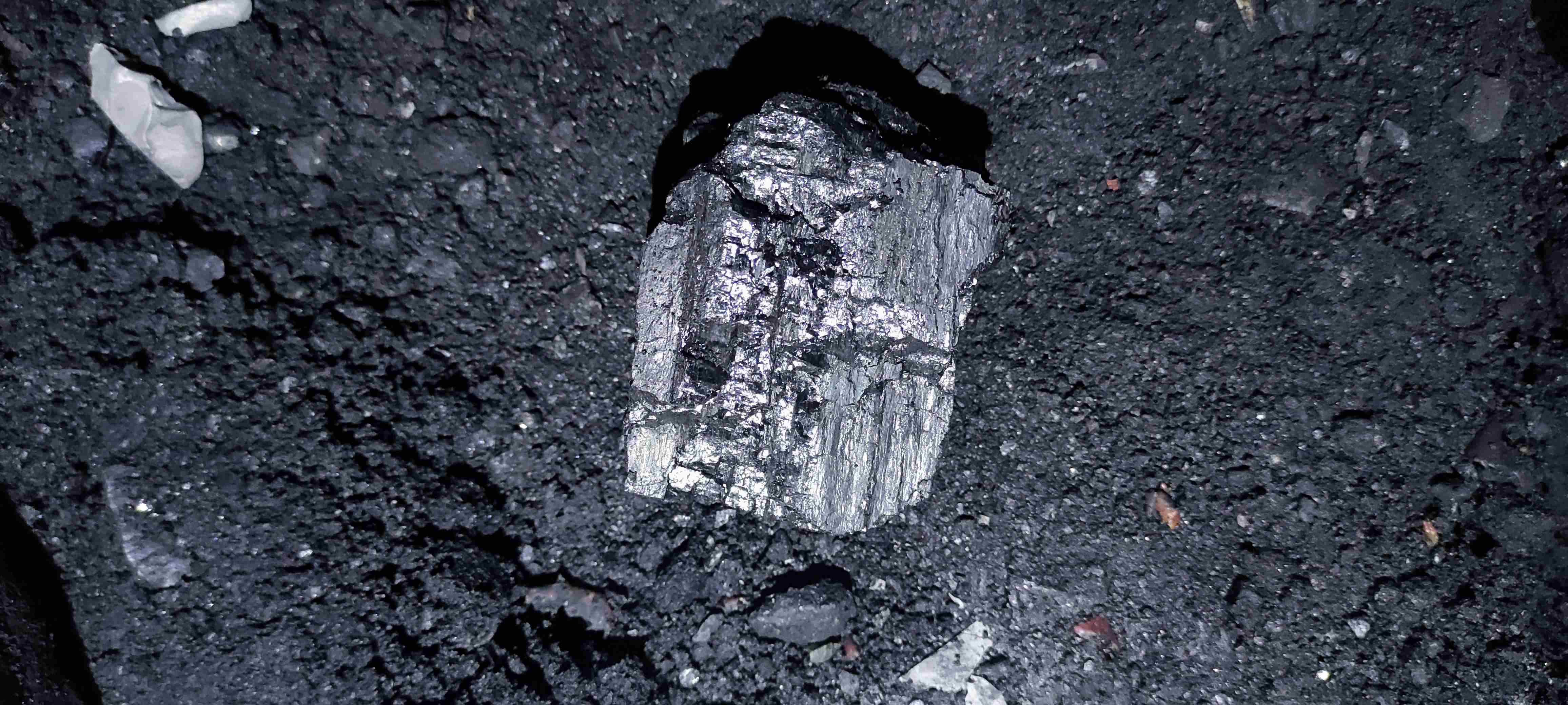}}
	\hfil
	\subfloat[]{\includegraphics[width=1.7in,height=1.5in]{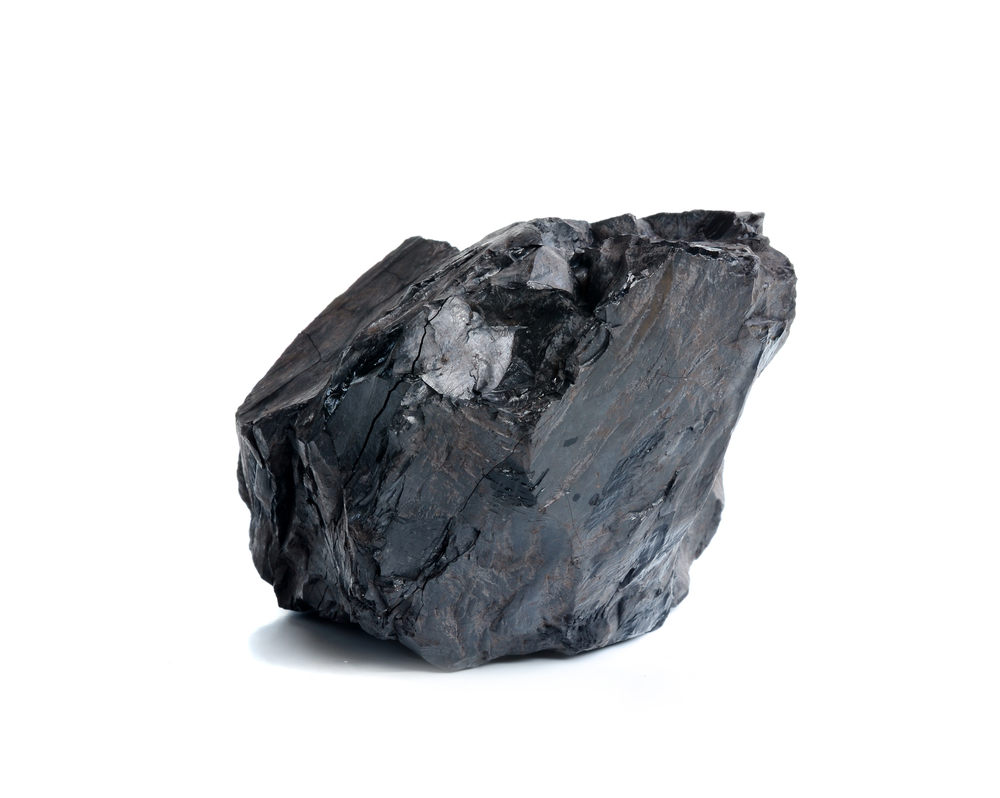}}
	
	\subfloat[]{\includegraphics[width=1.7in,height=1.5in]{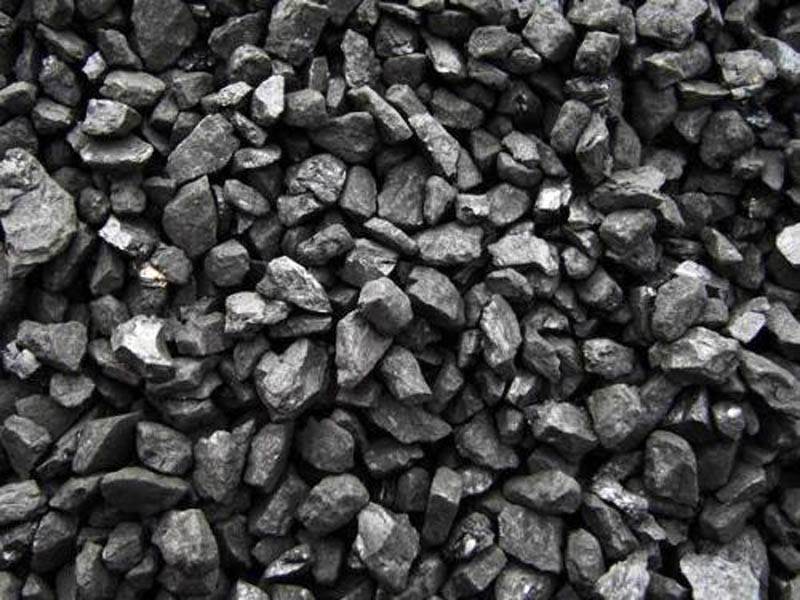}}
	\hfil
	\subfloat[]{\includegraphics[width=1.7in,height=1.5in]{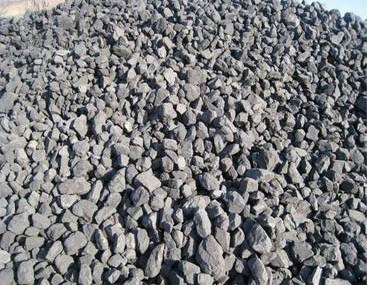}}
	\hfil
	\subfloat[]{\includegraphics[width=1.7in,height=1.5in]{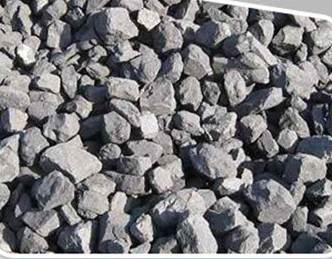}}
	\hfil
	\subfloat[]{\includegraphics[width=1.7in,height=1.5in]{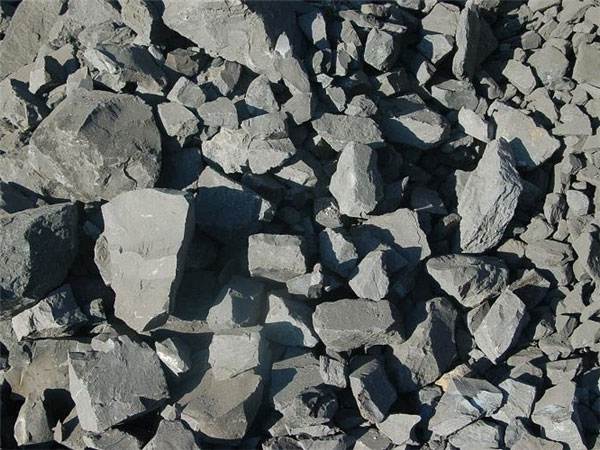}}

	\subfloat[]{\includegraphics[width=1.7in,height=1.5in]{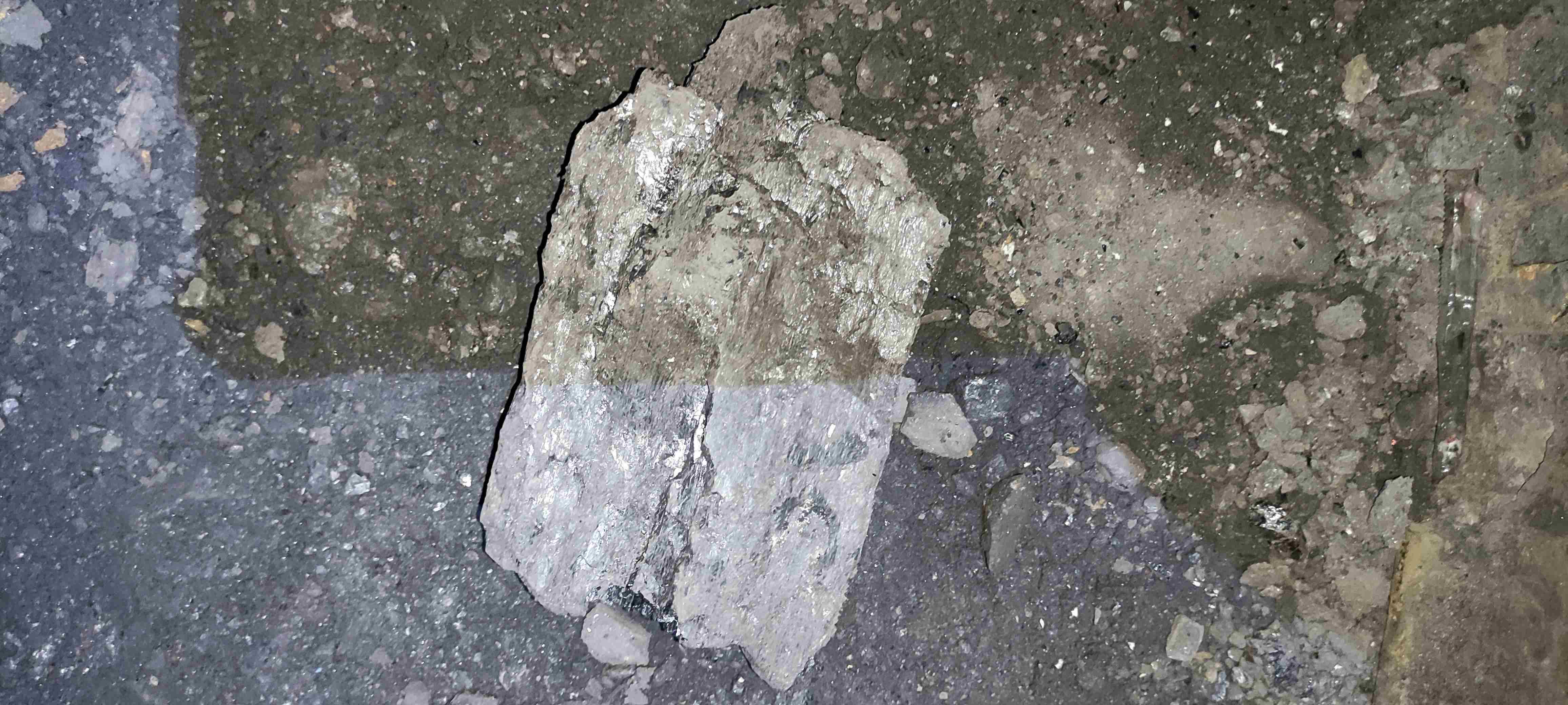}}
	\hfil
	\subfloat[]{\includegraphics[width=1.7in,height=1.5in]{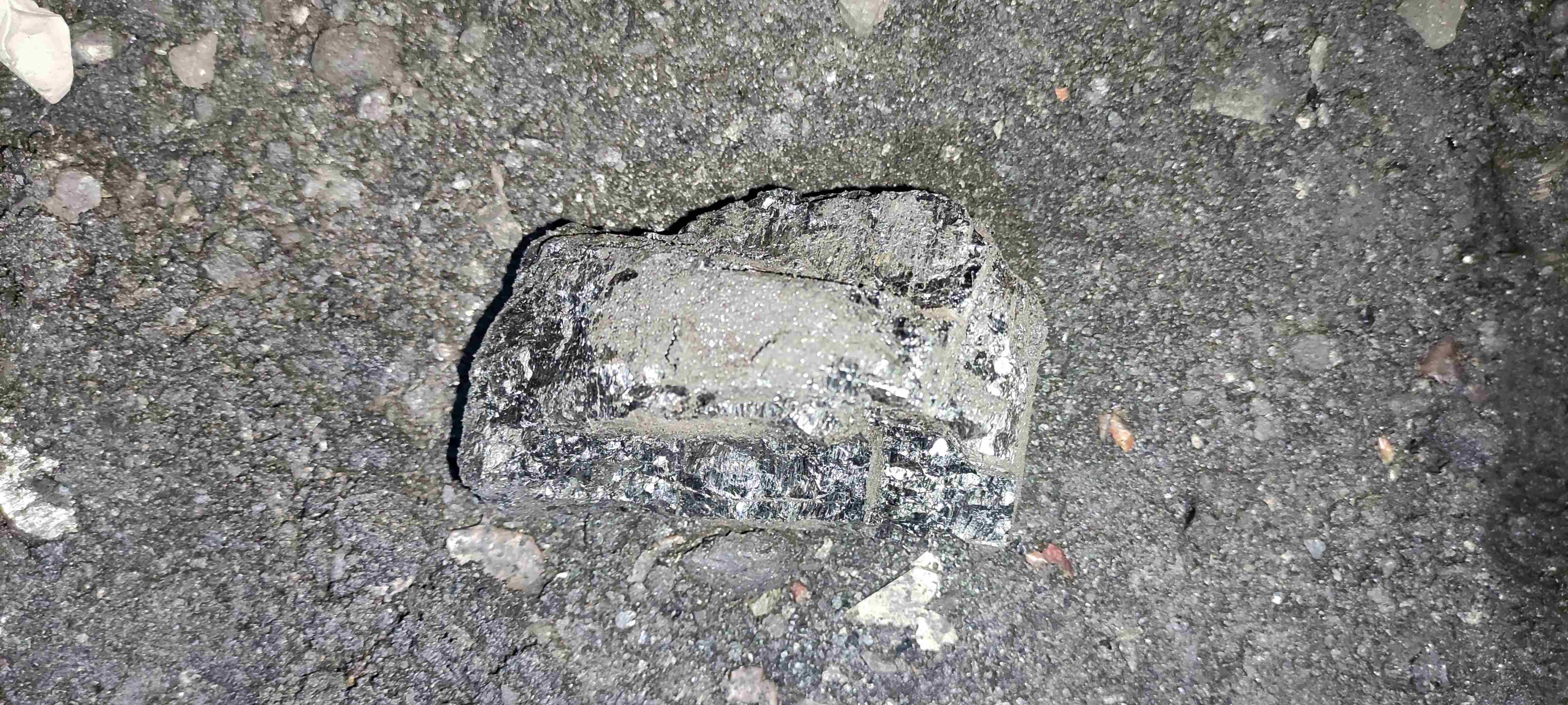}}
	\hfil
	\subfloat[]{\includegraphics[width=1.7in,height=1.5in]{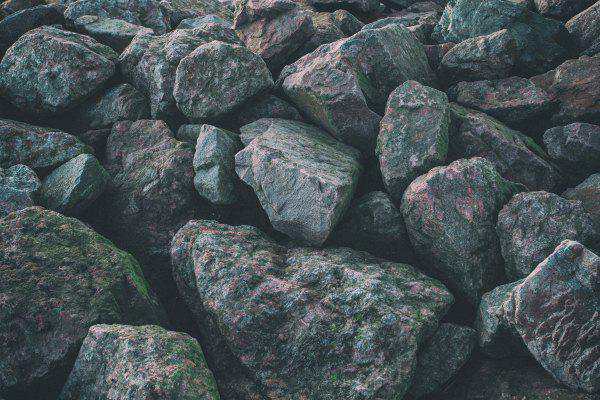}}
	\hfil
	\subfloat[]{\includegraphics[width=1.7in,height=1.5in]{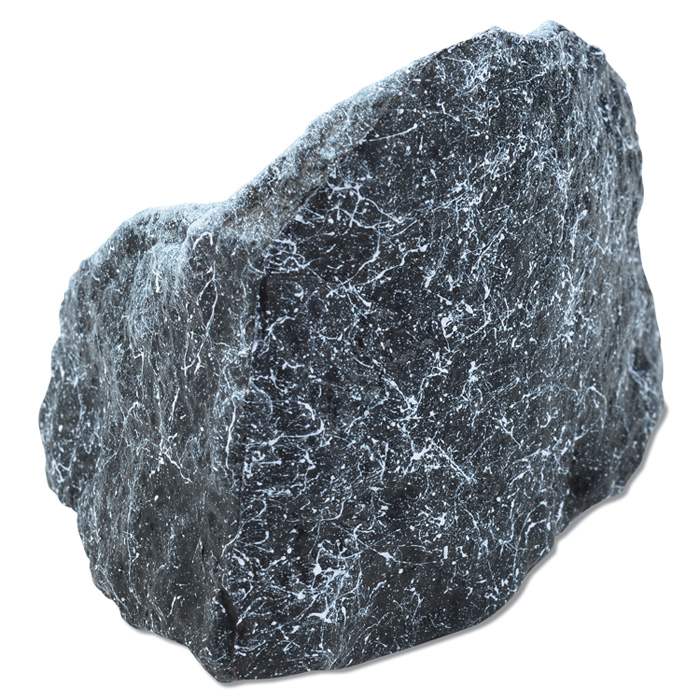}}
	\caption{Images from our datasets showing examples of coal, gangue and rock, where (a) (b) (c) (d) are coal, (e) (f) (g) (h) are gangue and (i) (j) (k) (l) are rock. (c) (i) (j) are real images taken under coal mine, we can clearly observe that they are in an environment with insufficient light conditions.}
	\label{dataset}
\end{figure*}
Our datasets consist of 212 coal and 686 gangue/rock images which acquired from Internet and labeled manually, along with 31 coal and 5 gangue images taken in Zhang Shuanglou Coal Mine located in Xuzhou, Jiangsu, China, shown in Fig.~\ref{dataset}. All coal images taken from mine are divided into test sets.
\subsection{Image augmentation}
Image augmentation is referred to making several changes to training images, thereby expanding the size of datasets, allowing models to learn more general characteristics rather than specific attributes, then avoid overfitting to some extent. In our experiments, images are resized to $300\times 300$, cropped at the center to $224\times 224$, randomly flipped horizontally and vertically then normalized using default ImageNet \cite{deng2009imagenet} mean and standard deviation:
\begin{equation}
\begin{aligned}
\text{MEAN}&=[0.485, 0.456, 0.406],\\
\text{STD}&=[0.229, 0.224, 0.225].
\end{aligned}
\end{equation}
\subsection{Training}
All the models (except SVM) are trained using Adam \cite{kingma2014adam}, and the \textit{smallest} variant in the original paper is used. We use batch size 128 and 50 epochs, respectively. The learning rate is set to 0.001, and a weight decay of 0.001 is used.

\textbf{LBP}. We calculate multi-scale local binary patterns of each image, including $\text{LBP}_{8,1}$, $\text{LBP}_{16,2}$, $\text{LBP}_{24,3}$ and $\text{LBP}_{24,4}$, where $\text{LBP}_{p,r}$ denotes the calculation of $p$ pixel points in a circle of radius $r$.

\textbf{GLCM}. For the gray-level co-occurrence matrix of each coal-rock image, several texture properties are calculated (including contrast, dissimilarity, homogeneity, energy, correlation and ASM) then concatenated as its feature.

\textbf{MSWS}. A 3-level wavelet decomposition is performed on each channel of the image using standard Haar filter. Only the high-frequency coefficient parts of the decomposition results are taken out and their means and variances are calculated as features. Hence, for each image, the dimension of feature is $3\times 2\times 3\times 3=54$.

\textbf{ResNet-18}. ResNet-18 is an improved CNN which introduces residual connections between convolutional layers \cite{he2016deep}. It consists of a convolutional layer with kernel size $7\times 7\times 64$ and stride 2, following by a max pooling layer with kernel size $3\times 3$ and stride 2, four residual-connected convolutional blocks and an average pooling layer. Each convolutional module has two convolutional layers with the same kernel size, and the kernel sizes are $3\times 3\times 64$,  $3\times 3\times 128$, $3\times 3\times 256$, $3\times 3\times 512$, respectively.

\textbf{ViT-B/16}. Vision Transformer is a relatively new model proposed by Google \cite{dosovitskiy2020image}. Its architecture can be considered as the encoder part of original Transformer \cite{vaswani2017attention}. The image $\mathbf{x}\in\mathbb{R}^{H\times W\times C}$ is reshaped into flattened 2D patches $\mathbf{x}_p\in\mathbb{R}^{N\times(P^2\cdot C)}$ ($P$: patch size, $N$: resulting number of patches) then projected to patch embeddings with dimension $D$. Multi-head self-attention (MSA) is calculated between patch embeddings along with a special learning token $\mathbf{x}_{\text{class}}$. Then, a multilayer perceptron (MLP) block with a hidden layer takes the result of MSA as input, followed by another MLP which used to output predicted probabilities for each class of the target. In our experiments, ViT-Base/16 is used, where 16 stands for patch size $P$. It has 12 layers of Transformer encoder with $D=768$, each contains one MSA with 12 heads and one MLP of size 3072.

\textbf{DenseNet-121}. After the input goes through a $7\times 7$ convolution and a pooling layer, the next 4 dense blocks and 3 transition layers are alternately connected, followed by a $7\times 7$ global average pooling layer and a MLP. In DenseNet-121, $N$ (see Sec. \ref{dense}) in 4 dense blocks is set to 6, 12, 24 and 16, respectively.

\textbf{DenseNet-GP}. Because we use DenseNet to extract features, then use Gaussian process to classify, we must ensure that features extracted is appropriate. Training DenseNet separately then fix its weights as an independent feature extractor is a solution, but in Sec. \ref{clf result} we will see that this will lead relatively low performance due to noises in labels. Hence, we use official implementation of DenseNet by PyTorch \cite{paszke2019pytorch} and Gaussian process by GPyTorch \cite{gardner2018gpytorch} then combine them to a single neural network.
\subsection{Classification results}
\label{clf result}
\begin{table}[!htb]
	\centering
	\caption{Classification results}
	\label{results}
	\begin{tabular}{c|c|c}
		\hhline
		Model           & Train Accuracy & Test Accuracy \\ \hline
		LBP + SVM         & 0.890 & 0.848 \\ \hline
		GLCM + SVM       & \textbf{1.000} & 0.770\\ \hline
		MSWS + SVM     & \textbf{1.000} & 0.804 \\ \hline
		ResNet-18      & 0.957 & 0.822 \\ \hline
		ViT-B/16       & 0.851 & 0.808 \\ \hline
		DenseNet-121         & 0.888 & 0.858 \\ \hline
		DenseNet-GP   & 0.900 & \textbf{0.899} \\
		\hhline
	\end{tabular}
\end{table}
\begin{figure*}[!t]
	\centering
	\subfloat[ResNet-18]{\includegraphics[width=1.7in]{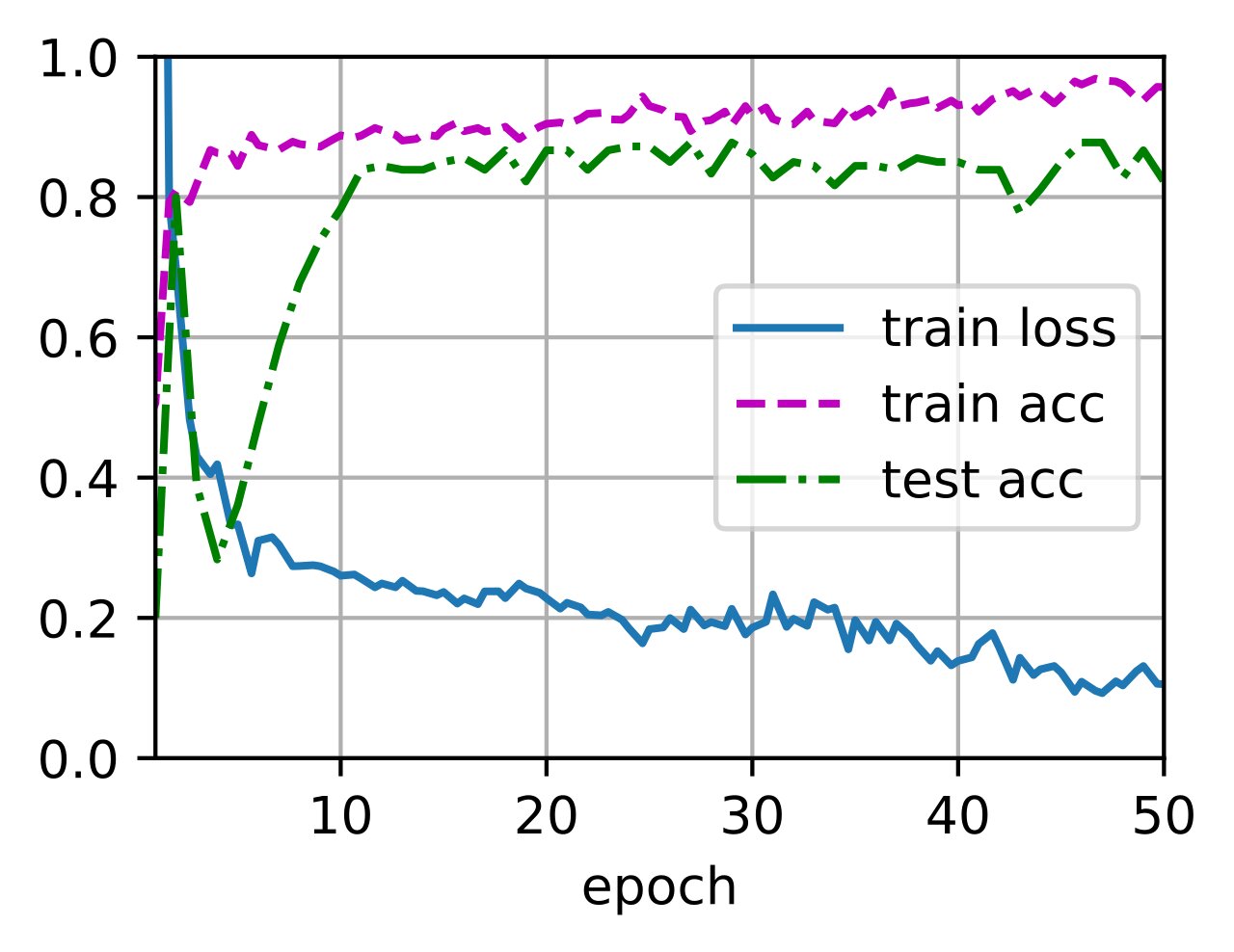}}
	\hfil
	\subfloat[ViT-B/16]{\includegraphics[width=1.7in]{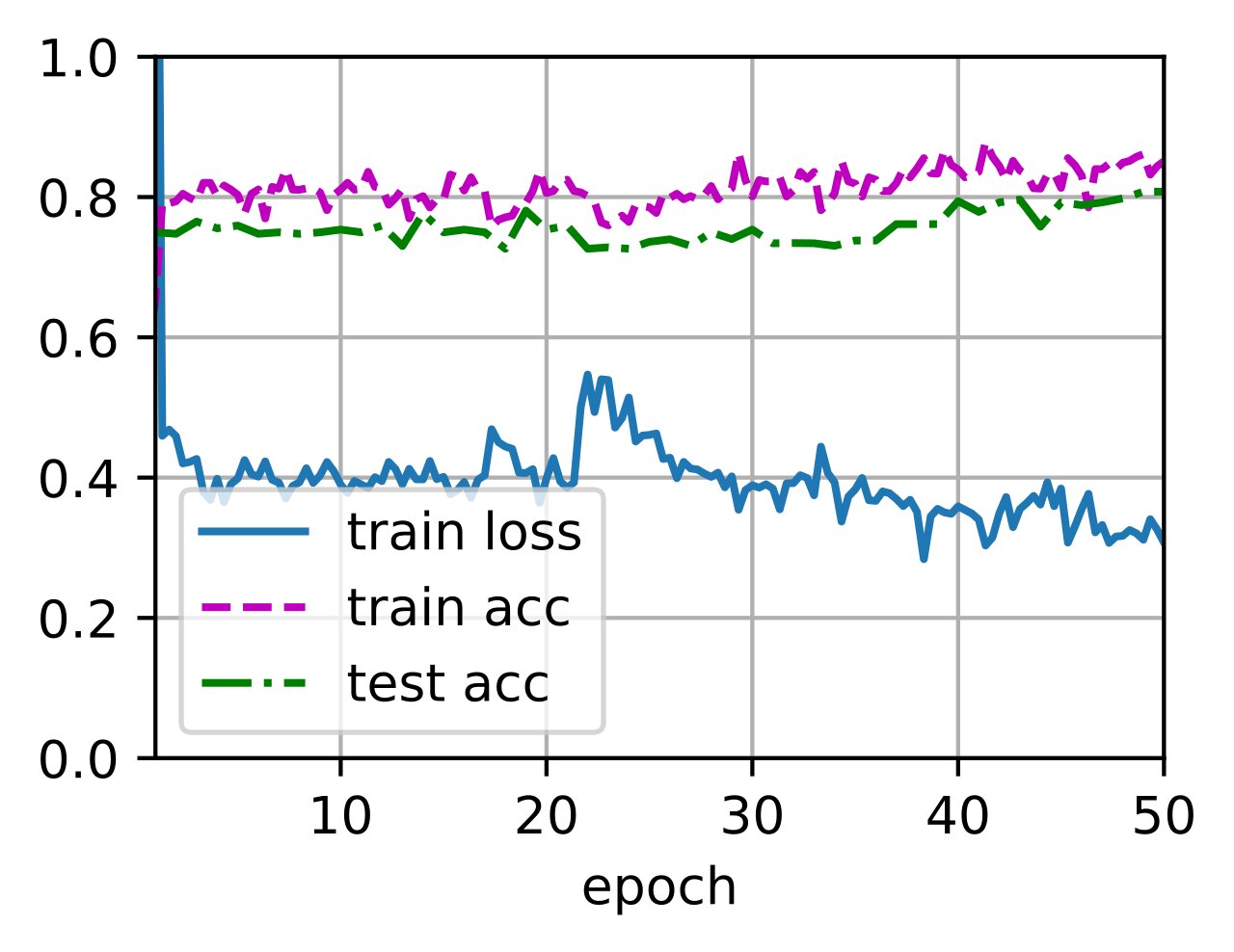}}
	\hfil
	\subfloat[DenseNet-121]{\includegraphics[width=1.7in]{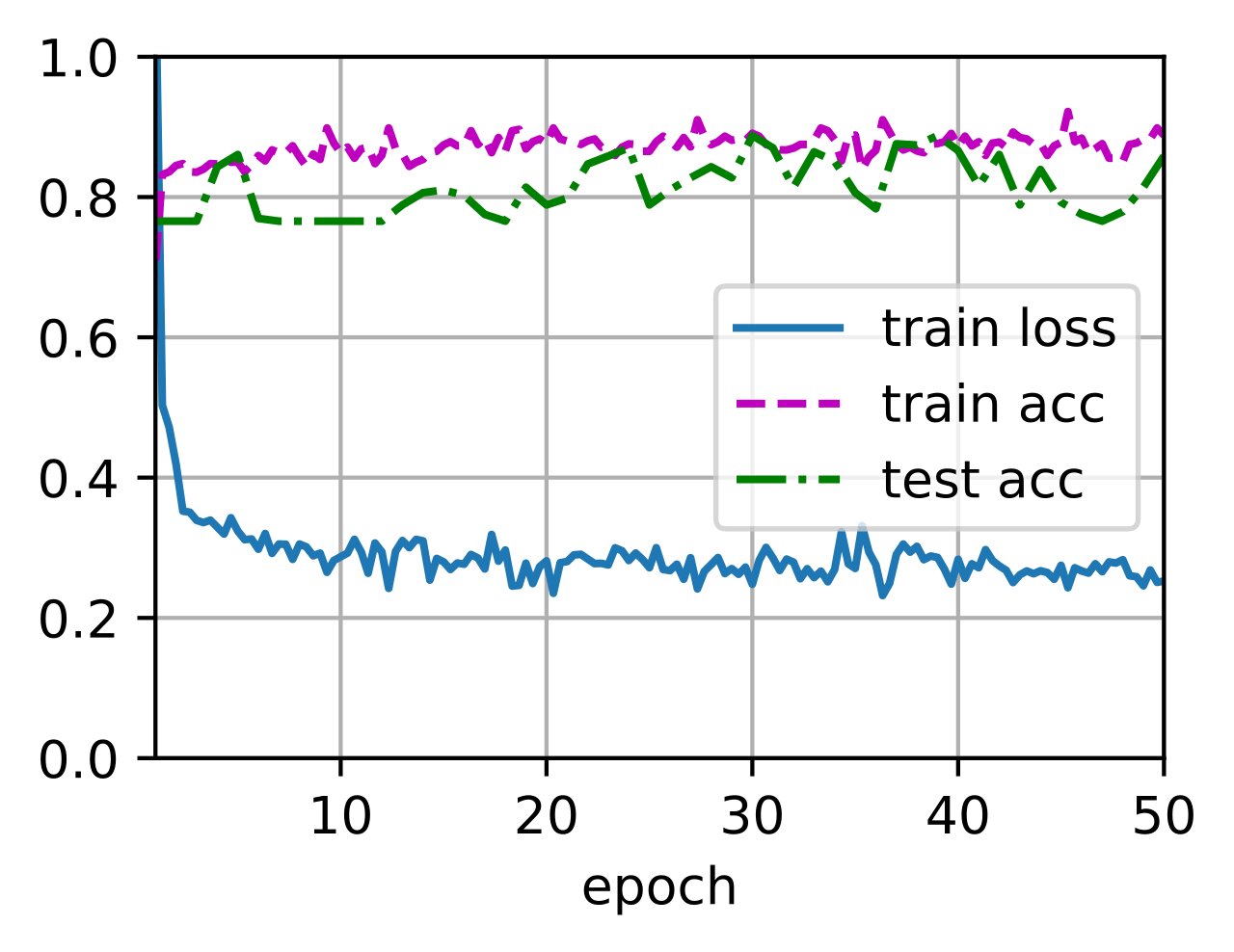}}
	\hfil
	\subfloat[DenseNet-GP]{\includegraphics[width=1.7in]{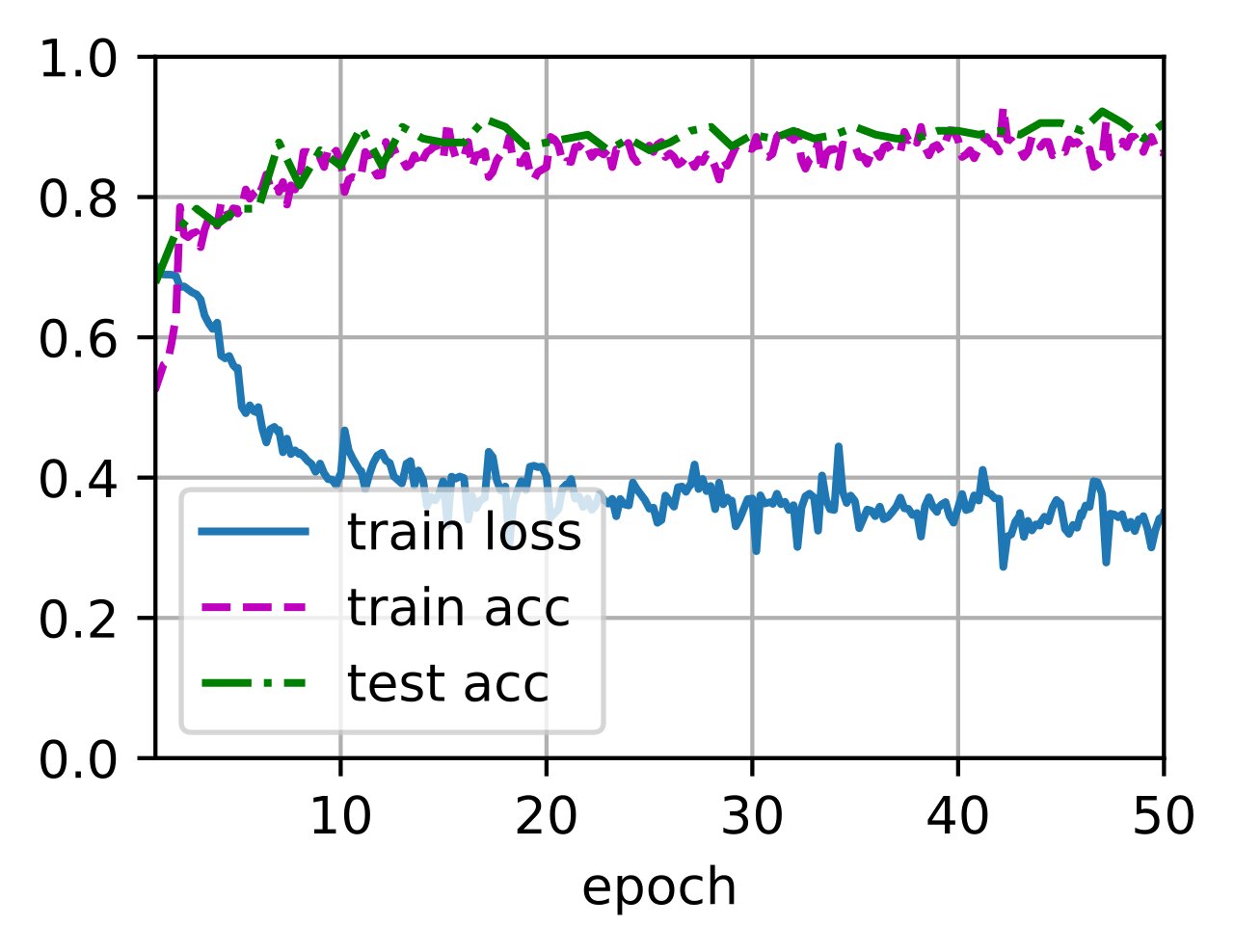}}
	\caption{Change curve of losses, train accuracies and test accuracies.}
	\label{loss}
\end{figure*}
Due to the groundtruth of few coal images crawled from Internet cannot be observed, there might be some noise in the label of training data. Our model outperforms other even in the presence of noise, see Table~\ref{results}.

\textbf{SVM}. The SVM method which uses GLCM and MSWS performs well on training data, perfectly separates coal from coal waste but achieves low test accuracy, which indicates SVM cannot identify noise in datasets or selected method to extract features cannot describe the whole distribution.

\textbf{Deep learning models}. Among conventional neural network models, severe overfitting occurs when using ResNet, yielding large gap between test accuracy and training accuracy. We also notice that ViT model even not converges on training datasets. Due to lack of image-specific inductive bias, ViT usually requires more data than CNNs to train. DenseNet-only model achieves higher test accuracy than ResNet, showing the advantages of dense connections over residual connections.

The losses and accuracies change during training process are recorded in Fig.~\ref{loss}. We can observe that test accuracy curves and training accuracy curves of ResNet-18 and DenseNet-121 are extremely incompatible, which supports our conclusion that they are overfitted. With regard to ViT-B/16 and DenseNet-GP, two curves are almost consistent, but ViT-B/16 does not converges due to lack of training data. From Fig.~\ref{loss}, we also discover that test accuracy of DenseNet-GP reaches 90\% only after training for 10 epochs. Hence, we draw the conclusion that DenseNet-GP achieves good results in terms of speed and accuracy.
\section{Conclusion}
We propose a new architecture which is called DenseNet-GP. It combines an improved convolutional neural network --- DenseNet and Gaussian process, perfectly absorbing the advantages of both. The DenseNet avoids overfitting, while Gaussian process handles noise in labels. Finally, our model achieves best accuracy, compared with other state-of-art models. Our model can distinguish coal images very well, even those that cannot be distinguished by the naked eye, while no other model works properly. In summary, our model has a certain reference value for realizing automatic identification and purification of coal mines by splitting coal waste and coal.

In the future we will try to validate the classification results in real working conditions. The data collected from the experiments are not all representative of the actual working environment, and the coal mining process is dynamic rather than static, so extending the classifier into a video workflow is also a challenge that needs to be addressed.

\bibliographystyle{unsrt}
\bibliography{ref}

\end{document}